\documentclass[twocolumn, 10pt]{article}

\usepackage{geometry}
\fontsize{12pt}{12pt} \selectfont

\usepackage[utf8]{inputenc} 
\usepackage[T1]{fontenc}    
\usepackage{hyperref}       
\usepackage{url}            
\usepackage{booktabs}       
\usepackage{amsfonts}       
\usepackage{nicefrac}       
\usepackage{microtype}      

\usepackage{amssymb,amsmath,amsthm,amsfonts}
\newtheorem{theorem}{Theorem}

\usepackage{algorithm}
\usepackage[noend]{algpseudocode}
\usepackage[pdftex]{graphicx}
\usepackage{multicol}
\newcommand{\formula}[1]{\begin{multline*} #1 \end{multline*}}
\newcommand{\clearpagebiblio}{\clearpage}
\newcommand{\dospace}{ \vspace{0.3cm} }
\newcommand{\insertplotsgbnips}{ }

\newcommand{\insertplotsforestnips}{ }

\newif\ifarxiv
\arxivtrue

\title{InfiniteBoost: building infinite ensembles with gradient descent}
\date{}
\author{
  \begin{tabular}[t]{cc}
    \parbox[t]{8cm}{\center {\bf Alex~Rogozhnikov}\footnote{Blog: http://arogozhnikov.github.io} \\
    Faculty of Mathematics \\
    National Research University \\ Higher School of Economics\\
    Moscow, Russia \\
    \texttt{alex.rogozhnikov@yandex.ru} }
    & \parbox[t]{8cm}{ \center
    {\bf Tatiana Likhomanenko} \\
    National Research Centre \\ "Kurchatov Institute" \\
    Moscow, Russia \\
    \texttt{tata.antares@yandex.ru} }
  \end{tabular}
}

\begin{document}

\maketitle

\begin{abstract}
    Ensemble methods have demonstrated high accuracy for a~variety of problems in different areas of machine learning.
    Two notable ensemble methods widely used in practice are gradient boosting and random forests.
    In this paper we present InfiniteBoost --- a novel algorithm, which combines important properties of these two approaches. 
    The algorithm constructs an ensemble of trees for which the following two properties hold: 
    trees of the ensemble account for mistakes of each other and at the same time the ensemble may contain an infinite number of trees without over-fitting.
    The proposed algorithm is evaluated on regression, classification, and ranking tasks using large scale, publicly available datasets.
\end{abstract}

\section{Introduction}

Nowadays ensemble methods of machine learning are one of the most widespread approaches used in many applications in industry and science:
search engines, recommendations, store sales prediction, customer behavior prediction, web text classification, high energy physics, astronomy, computational biology and chemistry, etc.
They achieve state-of-the-art results not only on many standard benchmarks~\cite{key-benchmarks,key-benchmarks-forest} but also in real-world tasks, like movements of individual body parts prediction~\cite{key-kinect}, click through rate prediction~\cite{key-facebook}, ranking relevance prediction in search engines~\cite{key-yahoo}. 
Wide applicability of ensembles is confirmed by data challenges: 60\% of winning solutions of challenges on Kaggle used XGBoost~\cite{key-xgboost}, one of the popular gradient boosting implementations. 

In ensemble methods multiple individual predictors are trained for the same problem and after that are combined in some manner.
Predictors in the ensemble can be constructed independently, like in bagging~\cite{key-bagging} and, in particular, random forests~\cite{key-random-forest}, or depending upon the performance of the previous models, like in gradient boosting~\cite{key-gb}.

Random forests suggested by L. Breiman~\cite{key-random-forest} are a combination of tree predictors such that each tree in the ensemble is constructed uses a random subset of features and a bootstrap replica of the training set provided by bagging~\cite{key-bagging}.
Predictions of trees are aggregated by simple averaging.
It can be shown that generalization error for forests converges a.s. to a limit as the number of trees in the forest tends to infinity.
As a result, one can include arbitrarily many trees in the ensemble without decreasing its performance on unseen data.
Also L. Breiman shows that predictors in the ensemble should be accurate but uncorrelated to achieve a better quality than individual predictors. 
In practice accuracy of predictors is provided by employing deep trees, whereas low correlation is achieved by bagging and random features subsampling.

In contrast to random forests, a general gradient descent boosting paradigm~\cite{key-gb,key-sgb}, developed by J. Friedman, 
introduces consecutive ensemble building by greedily approximating a target function with gradient descent in the space of functions. 
On each iteration of boosting a new tree is constructed to approximate a gradient of the loss function, 
which provides accounting for the performance and mistakes of previously constructed trees in the ensemble.
In real world applications gradient boosting models often have to contain thousands of trees to provide the best possible quality.
However, building arbitrarily large boosting models drives to decrease of quality on the unseen data~\cite{key-brownboost, key-dart}, an effect known as over-fitting.

To fight mentioned disadvantages of random forest and gradient boosting we present a novel algorithm, called InfiniteBoost, the goal of which is to combine the best properties of both: allow for construction of an infinite (arbitrarily large) ensemble and at the same time account for the mistakes of previously constructed trees in the ensemble using a gradient descent method.

The rest of this paper is organized as follows: in Section~2 we explain the motivation behind InfiniteBoost; in Section~3 we describe InfiniteBoost algorithm and prove the convergence theorem.
Comparison of the proposed algorithm with gradient boosting and random forests is given in Section~4: experimental results are listed for regression, classification, and ranking problems. 
Finally, we provide an overview of other existing boosting-bagging hybrid algorithms (term introduced by J. Friedman~\cite{key-sgb}) in Section~5.

\section{Motivation}

For simplicity we consider specifically ensembles over decision trees, while proposed framework naturally generalizes to ensembling of arbitrary models.
Random forests construct trees in the ensemble independently using randomness in feature selection and bagging to provide own training set for each predictor.
This independence makes it possible to build infinite ensembles: quality on the unseen data converges to some constant as the number of trees in the ensemble becomes large.

In contrast to random forests, gradient boosting procedure allows next tree to account for the mistakes done by the previous trees in the ensemble.
However, it is known (and often seen in applications) that gradient boosting is prone to over-fitting.
In practice over-fitting can be detected using the learning curve (quality vs number of boosting iterations) when quality on the holdout increases depending on the iteration of boosting and then decreases starting from some iteration. 
Thus, for gradient boosting an arbitrarily large number of predictors in an ensemble is not an option. 


To reduce over-fitting effect and allow construction of larger ensembles, a shrinkage parameter is introduced to the boosting procedure: a new tree is added to the ensemble with a coefficient $\eta$, called shrinkage (or learning rate).
Shrinkage is an important hyperparameter of gradient boosting which requires careful tuning in applications.

InfiniteBoost is an algorithm which aims to build an infinite ensemble (without over-fitting effect with increasing the number of predictors) such that each new predictor incorporates the errors made by the previous predictors in the ensemble (boosting procedure).
We also demonstrate that tuning of shrinkage is not required for InfiniteBoost. 

\section{InfiniteBoost}

We start from analyzing the desired properties of an algorithm:
a) build an infinite converging ensemble 
b) account for the errors made by trees using gradient boosting approach.
Firstly, let us consider a plain gradient boosting algorithm (Algorithm~\ref{alg:gb}),
where the ensemble prediction $F(x)$ accumulates contribution of the trees $ F(x) = \eta \sum_{m=1}^M \text{tree}_m(x) $.

\begin{algorithm}[!h]
  \caption{Gradient boosting}\label{alg:gb}
  {\bf Input}: training set $\{x_i, y_i\}_{i=1}^n$; a differentiable loss function $L(y, F(x))$; number of boosting iterations $M$; shrinkage~$\eta$.

  {\bf Algorithm} ($F(x)$ is output):
  \begin{algorithmic}
    \State Initialize model with a zero value: $F(x) \gets 0$.
    \For{$m=1,\dots,M$}
      \State $\text{tree}_m \gets \text{learn}\left(\left\{x_i, -\frac{\partial L(y_i, F(x_i))}{\partial F(x_i)}\right\}_{i=1}^n\right)$
      \State $F(x) \gets F(x) + \eta \text{ tree}_m(x)$
    \EndFor
  \end{algorithmic}

  Optional steps like finding initial constant prediction, usage of Newton-Raphson step,
  or usage of Hessian~\cite{key-xgboost} in the tree construction are skipped for brevity.
\end{algorithm}

To achieve a stationary state (condition a)), individual predictions in InfiniteBoost are averaged with weights $\alpha_m$
(that can be taken uniform $\alpha_m = 1$)
and scaled by an additional constant $c$, called ensemble capacity (or simply capacity):
$ F(x) = c \times \dfrac{\sum_m \alpha_m \text{tree}_m(x)}{ \sum_m \alpha_m } $.
The contribution of each individual tree in this model converges to zero.
To satisfy property a) an ensembling model should after sufficiently large amount of iterations converge to a stationary process,
and the distribution\footnote{We expect the process of tree building to be randomized, which is crucial in building powerful ensembles of trees} of newly-built trees should coincide with the distribution of trees already in the ensemble property)
\begin{equation*}
  F(x) = c \times \mathbb{E}_{\text{trees at }F(x)} \text{tree}(x)
\end{equation*}
with average taken over trees generated by usual training procedure
\formula{
    \text{trees at }F(x)= \\
    =\left\{\text{tree} \, \Big| \, \text{tree} \gets \text{learn}\left(\left\{x_i, -\frac{\partial L(y_i, F(x_i))}{\partial F(x_i)}\right\}_{i=1}^n\right)\right\}.
}
The building process is determined by a vector $z$ of the current ensemble's predictions on the training set:
$ z = ( F(x_1), \dots, F(x_n) ) $.
Let $ \overline{T}(z)_i = \mathbb{E}_\text{tree at $z$} \text{tree}(x_i) $ be the average prediction of an ensemble on the training set. Then we can see that the stationary point of the process is a solution of an equation
\begin{equation}
    z = c \times \overline{T}(z) \label{eq:stationary}.
\end{equation}

\begin{theorem}
Equation~(\ref{eq:stationary}) has a solution if  
$\overline{T}(z)$ is bounded ($|| \overline{T}(z) || < \text{const}$) and continuous.
\end{theorem}
This follows directly from Brouwer fixed-point theorem.
Boundedness holds if the gradient is bounded (as in logistic loss),
for other losses this can be achieved by scaling the gradient after some (arbitrarily large) threshold.
As for continuity of $\overline{T}(z)$,
this property can be enforced e.g. by adding Gaussian noise to $z$ before building a tree.

\begin{algorithm}[!h]
  \caption{Infinite Boosting (InifiniteBoost)}\label{alg:infiniteboost}
  {\bf Input}: training set $\{x_i, y_i\}_{i=1}^n$; a differentiable loss function $L(y, F(x))$; the number of boosting iterations $M$; capacity $c$.

  {\bf Algorithm} ($F(x)$ is output):
  \begin{algorithmic}
    \State Initialize model with a zero value: $F(x) \gets 0$.
    \For{$m=1,\dots,M$}
      \State $\text{tree}_m \gets \text{learn}\left(\left\{x_i, -\frac{\partial L(y_i, F(x_i))}{\partial F(x_i)}\right\}_{i=1}^n\right)$
      \State $F(x) \gets  \frac{\sum_{k=1}^m \alpha_k \text{tree}_k(x)}{\sum_{k=1}^m \alpha_k} \times c$
    \EndFor
  \end{algorithmic}

  {\bf Remarks}:
  \begin{itemize}
    \item To avoid over-stepping in first iterations capacity $c$ is replaced with a value providing the contribution of a single tree not greater than one.
    \item To avoid recomputing ensemble ($\eta_m = \frac{\alpha_m}{\sum_{k=1}^m \alpha_k}$):
      \begin{equation*}
        F(x) \gets (1-\eta_m)F(x) + \eta_m \times c \times \text{tree}_m(x).
      \end{equation*}
  \end{itemize}
\end{algorithm}
A method we propose to find this stationary state is InfiniteBoost (Algorithm~\ref{alg:infiniteboost}).
It can be interpreted as a stochastic optimization of the following regularized loss function:
\begin{equation*}
  \dfrac{|| z ||^2}{2} + c \sum_{i=1}^n L(y_i, z_i) = \dfrac{|| z ||^2}{2} + c \mathcal{L},
\end{equation*}
with stochastic gradient descent update rule:
\formula{
    z \leftarrow z - \eta_m \left(z + c \times \text{grad}_z \mathcal{L} \right) = \\
    = (1 - \eta_m) z - \eta_m c \times \text{grad}_z \mathcal{L}.
}
As in usual gradient boosting, a gradient step is replaced with building a tree modelling a negative gradient.
Then an ensemble prediction is updated similarly:
\begin{equation*}
    F(x) \leftarrow (1 - \eta_m) F(x) + \eta_m c \times \text{tree}_m(x).
\end{equation*}
Different choices of $\eta_m$ are possible and correspond to different weightings $\alpha_m$.
For example, $\eta_m = 1 / m$ correspond to uniform weighting $\alpha_m = 1$,
while $\eta_m = 2 / (m + 1)$ correspond to assigning higher weights for later trees $\alpha_m = m$.
In our experiments the second option is used, because it keeps the effective sample size of $\frac{3}{4}m$ (compatible with uniform weighting),
but allows ensemble to adapt faster to new values of $z$:
to obtain 99\% contribution of "new trees" to the ensemble one needs to enlarge size of the ensemble by a factor of 10,
whereas with uniform weighting the size should be enlarged by 100 times.



\begin{theorem}
InfiniteBoost converges almost surely to the solution of Equation (\ref{eq:stationary}),
provided that $c \times \overline{T}(z)$ is a contraction mapping and
$\eta_m \sim \frac{1}{m}$.
\end{theorem}

The contraction requirement may hold only for sufficiently small capacities $c$,
also it implies there is one and only one solution $z^{*}$ of Equation (\ref{eq:stationary}).
Theorem follows from an inequality
\formula{
    \mathbb{E} || z_{m+1} - z^{*} ||^2 \leq \\
    \leq \left(1 - \text{const}_1 \, \eta_m  \right)^2 \mathbb{E} || z_m - z^{*} ||^2
    + \text{const}_2 \, \eta^2_m
}
that holds for some positive constants $\text{const}_1$, $\text{const}_2$,
consequently, $\lim_{m \to +\infty} \mathbb{E} || z_{m} - z^{*} ||^2 = 0$.


\subsection{Adapting ensemble capacity during traning}

Capacity $c$ of an ensemble can be changed in the process of building, allowing next trees to find another optimal point.
Since the loss is differentiable with respect to capacity $c$, we present a variant of InfiniteBoost (Algorithm \ref{alg:infiniteboost_adaptive}) which adapts $c$ using holdout.
In experiments, 5\% of the training set is used as a holdout to tune capacity.
This way, one may skip the process of tuning capacity (or similar parameter shrinkage of gradient boosting).
After properly selecting capacity, holdout can be added to the training set to improve results, but this was not done in our experiments.

\begin{algorithm}[!h]
  \caption{InifiniteBoost with adaptive capacity}\label{alg:infiniteboost_adaptive}
  {\bf Input}: training set $\{x_i, y_i\}_{i=1}^n$; a differentiable loss function $L(y, F(x))$; the number of boosting iterations $M$.

  {\bf Algorithm} ($F(x)$ is output):
  \begin{algorithmic}
    \State Initialize: model $F(x) \gets 0$; capacity $c=\frac{1}{2}$.
    \State Divide training set into two non-overlapping subsets $\{x_i, y_i\}_{i=1}^n=T\sqcup H$.
    \For{$m=1,\dots,M$}
      \State $\text{tree}_m \gets \text{learn}\left(\left\{x_i, -\frac{\partial L(y_i, F(x_i))}{\partial F(x_i)}\right\}_{\{x_i, y_i\}\in T}\right)$
      \State $F(x) \gets \frac{\sum_{k=1}^m \alpha_k \text{tree}_k(x)}{\sum_{k=1}^m \alpha_k} \times c$
      \State Use very minor correction of capacity:
      \State $s \gets \text{sgn} \sum_{\{x_i, y_i\}\in H} -\frac{\partial L(y_i, F(x_i))}{\partial F(x_i)} F(x_i)$
      \State $c \gets c\times \left(\frac{m+1}{m}\right)^s$
    \EndFor
  \end{algorithmic}
\end{algorithm}

\section{Experiments}
We have evaluated InfiniteBoost for three different tasks: classification, regression, and ranking. 
Large scale, publicly available datasets are used in comparison. 
In our evaluation, we compare InfiniteBoost with two extreme cases: random forests and gradient boosting with different shrinkage values.\footnote{
  Reference implementation, code of all the experiments, and additional plots are available on the github: \url{https://github.com/arogozhnikov/infiniteboost}.
}

 \begin{table*}[!t]
  \caption{Datasets description for infinite boosting and gradient boosting comparison}
  \label{tab:gb-data}
  \centering
  \dospace
  \begin{tabular}{lllll}
    \toprule
    Type     &  Name & Number of instances     & Number of features & Source \\
    \midrule
    classification & Higgs 1M & 1,500,000  & 28   & \href{https://archive.ics.uci.edu/ml/datasets/HIGGS}{\underline{link}}  \\
    regression     & YearPredictionMSD & 515,345 & 90   & \href{https://archive.ics.uci.edu/ml/datasets/YearPredictionMSD}{\underline{link}}   \\
    ranking     & yahoo-letor, set 1 &  638,794   & 699  & \cite{key-yahoo-data} \\
    \bottomrule
  \end{tabular}
\end{table*}

\begin{figure*}[!h]
  \centering
  \begin{multicols}{2}
    \includegraphics[width=1\linewidth]{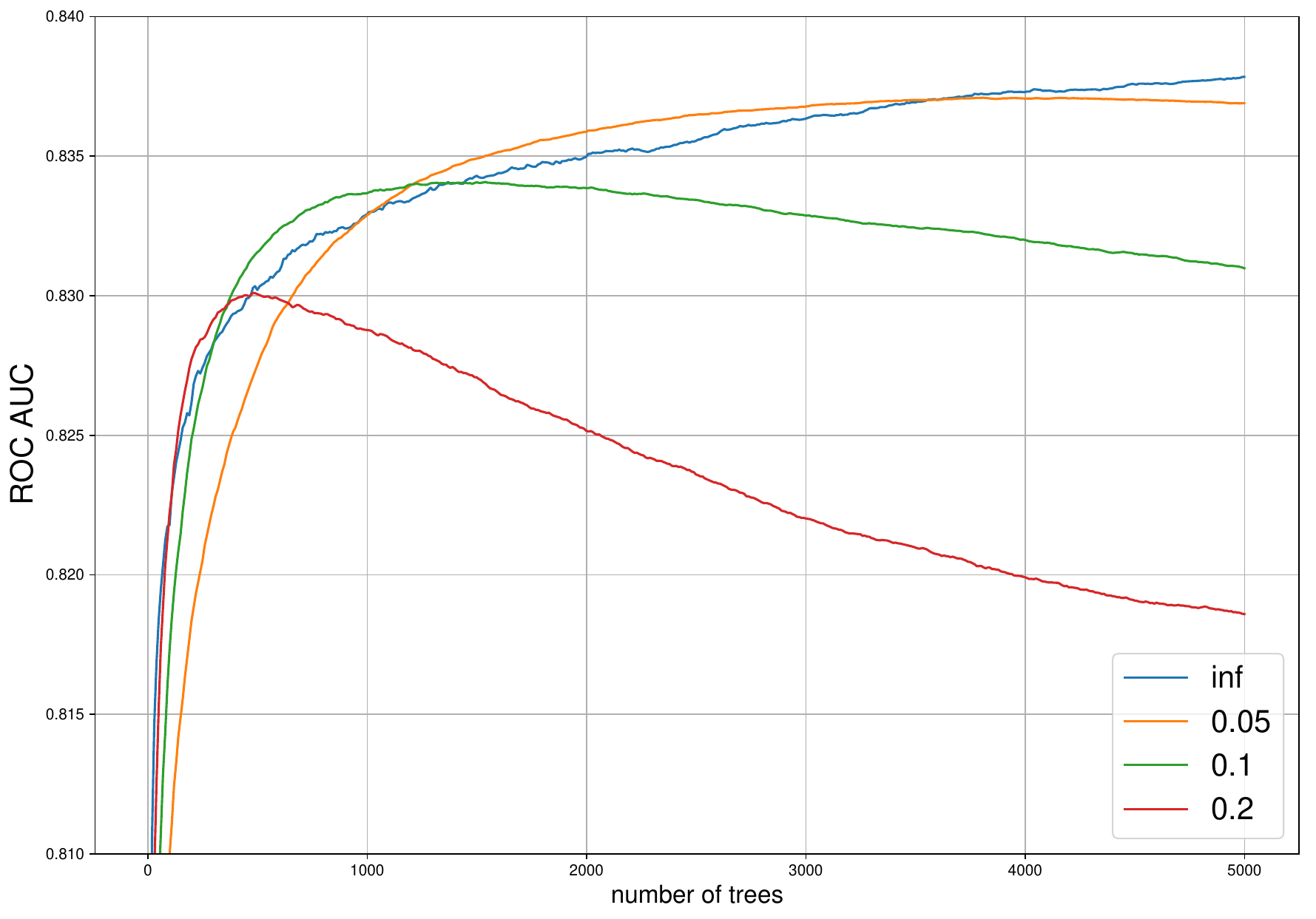}
    \caption{Quality on Higgs dataset for gradient boosting with different shrinkages ($0.05, \, 0.1,\, 0.2$) and infinite boosting with adaptive capacity (inf). \label{fig:gb-auc}}
    \includegraphics[width=1\linewidth]{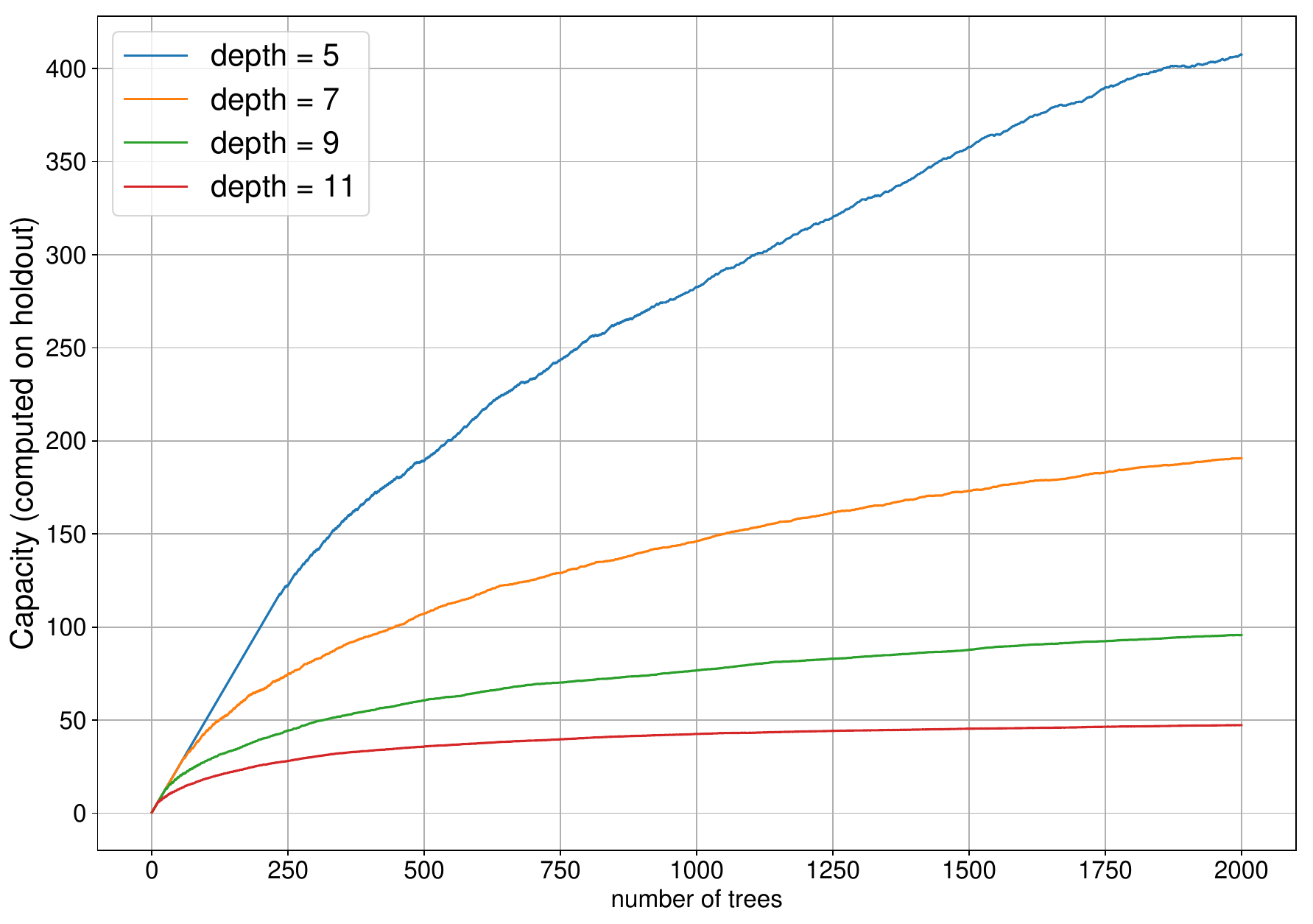}
    \caption{Comparison of infinite boost capacities found by adapting capacity on a holdout for different depths on Higgs dataset. \label{fig:gb-depth}}
  \end{multicols}
\end{figure*}

\begin{figure*}[!h]
  \centering
  \begin{multicols}{2}
    \includegraphics[width=1\linewidth]{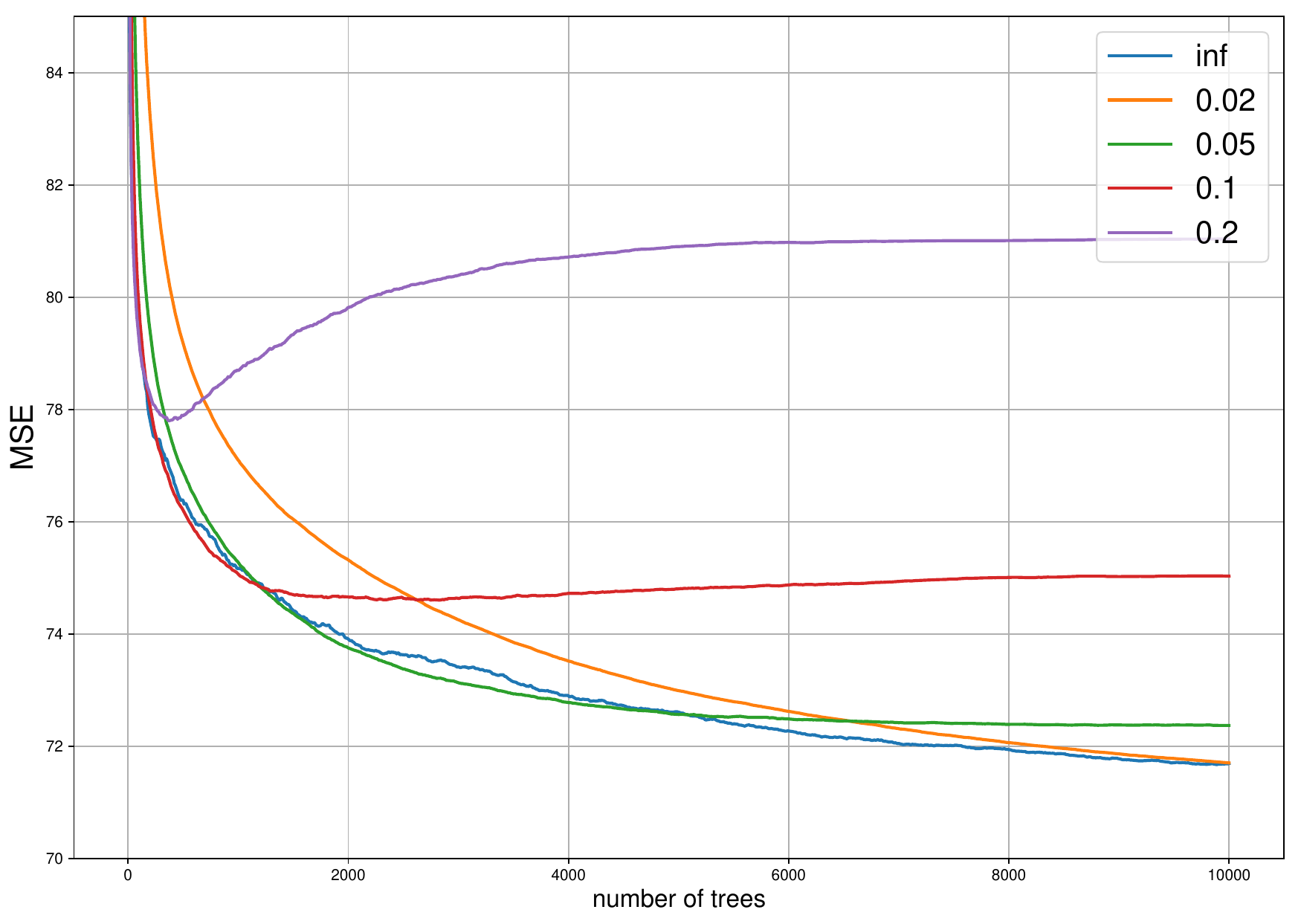}
    \caption{Quality on YearPredictionMSD dataset for gradient boosting with different shrinkages ($0.02,\, 0.05, \, 0.1,\, 0.2$) and infinite boosting with adaptive capacity (inf). \label{fig:gb-mse}}
    \includegraphics[width=1\linewidth]{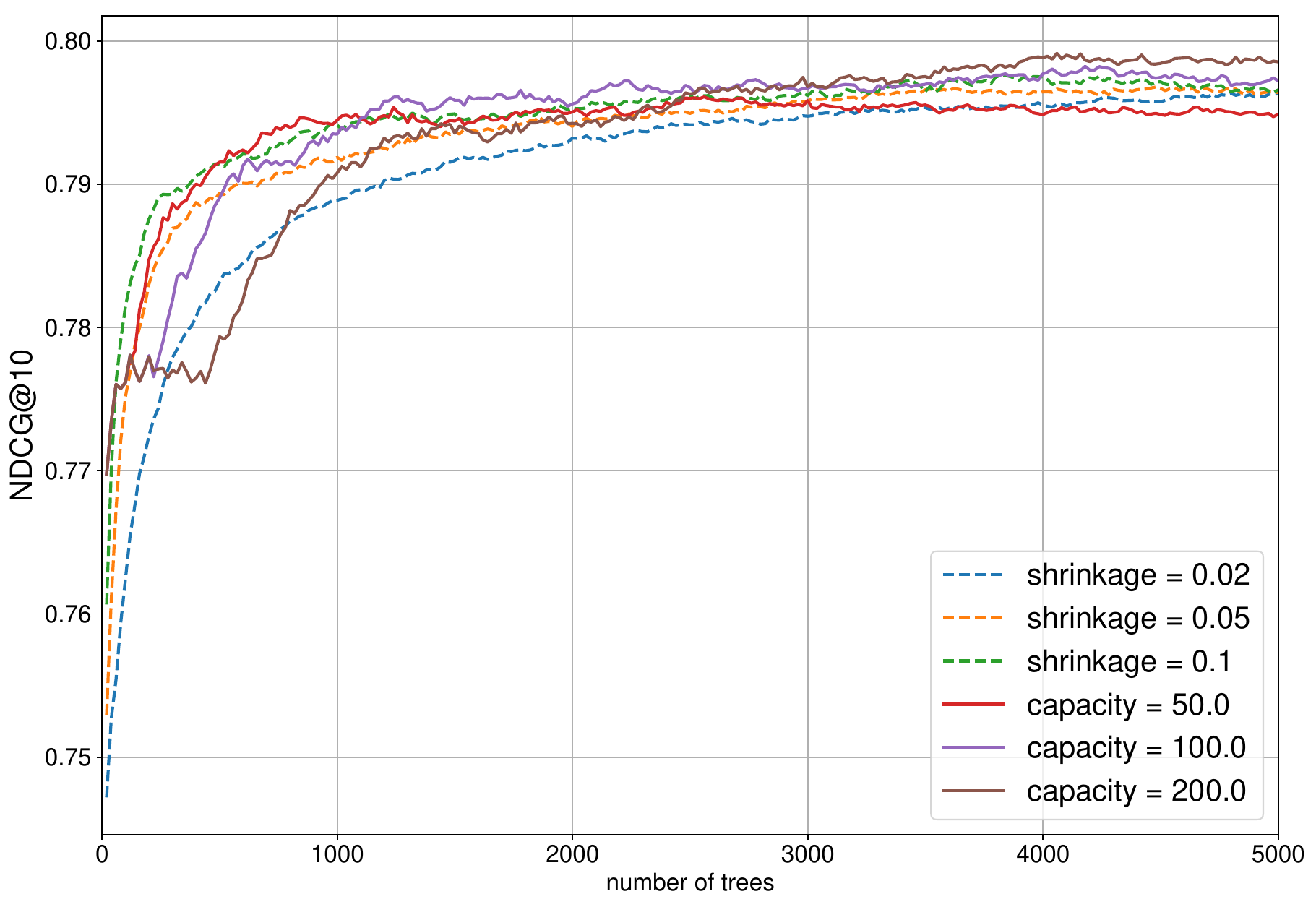}
    \caption{Quality on yahoo-letor dataset for gradient boosting with different shrinkages ($0.05, \, 0.1,\, 0.2$) and infinite boosting with different capacity values ($50,\, 100,\, 200$).
             Other NDCG@k plots can be found in supplementary material. \label{fig:gb-ndcg}}
  \end{multicols}
\end{figure*}

 \begin{table*}[!t]
  \caption{Datasets description for infinite boosting and random forest comparison}
  \label{tab:rf-data}
  \centering
  \dospace
  \begin{tabular}{lllll}
    \toprule
    Type     &  Name & Number of instances     & Number of features & Source \\
    \midrule
    classification & covertype & 581,012  & 54  & \href{https://archive.ics.uci.edu/ml/datasets/covertype}{\underline{link}}   \\
    classification     & real-sim  & 72,309 & 20,958  & \href{https://www.csie.ntu.edu.tw/~cjlin/libsvmtools/datasets/binary.html}{\underline{link}}     \\
    classification     & citeseer &  181,395 & 105,354 & \href{http://komarix.org/ac/ds/}{\underline{link}}\\
    \bottomrule
  \end{tabular}
\end{table*}

\begin{table*}[!t]
    \caption{
    ROC AUC qualities provided by random forest and infinite boosting with different capacities $c$ for classification tasks: ensembles contain 100 trees;
    provided values are the mean and the error computed with $k$-folding ($k=4$)
    }
    \label{tab:rf-auc}
    \centering
    \dospace
    \begin{tabular}{llll}
    \toprule
    {} &               covertype &               real-sim &              citeseer \\
    \midrule
    Random Forest      &  $ 0.9933 \pm 0.0001 $ &  $ 0.9907 \pm 0.0005 $ &  $ 0.8831 \pm 0.0086 $ \\
    InfiniteBoost, $c=1$ &  $ 0.9937 \pm 0.0000 $ &  $ 0.9914 \pm 0.0005 $ &  $ 0.8763 \pm 0.0132 $ \\
    InfiniteBoost, $c=2$ &  $ 0.9940 \pm 0.0001 $ &  $ 0.9918 \pm 0.0006 $ &  $ 0.8797 \pm 0.0160 $ \\
    InfiniteBoost, $c=4$ &  $ 0.9945 \pm 0.0001 $ &  $ 0.9931 \pm 0.0004 $ &  $ 0.8764 \pm 0.0172 $ \\
    \bottomrule
    \end{tabular}

\end{table*}

\ifarxiv
  \begin{figure*}[!h]
    \centering
    \begin{multicols}{2}
      \includegraphics[width=1\linewidth]{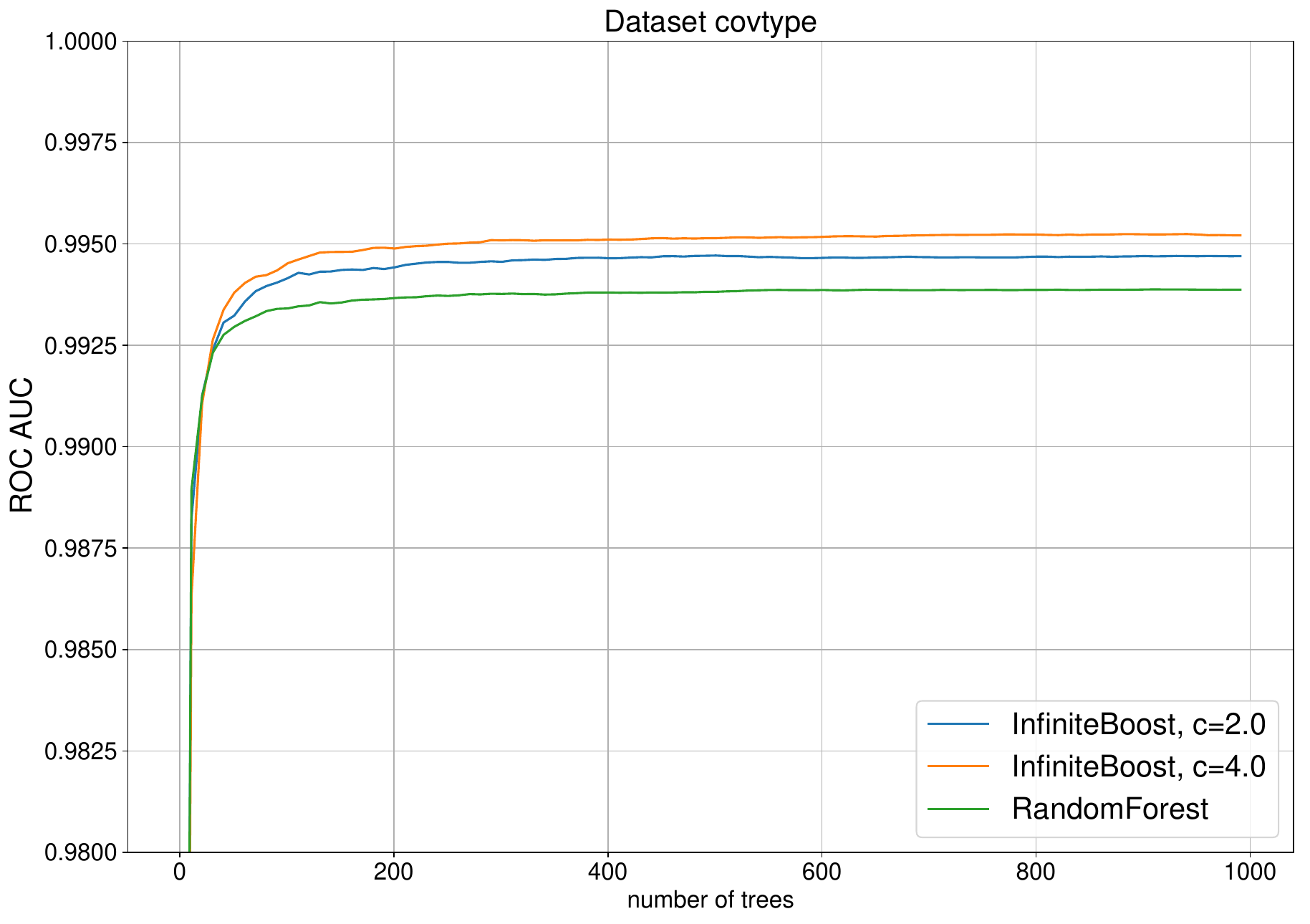}
      
      \includegraphics[width=1\linewidth]{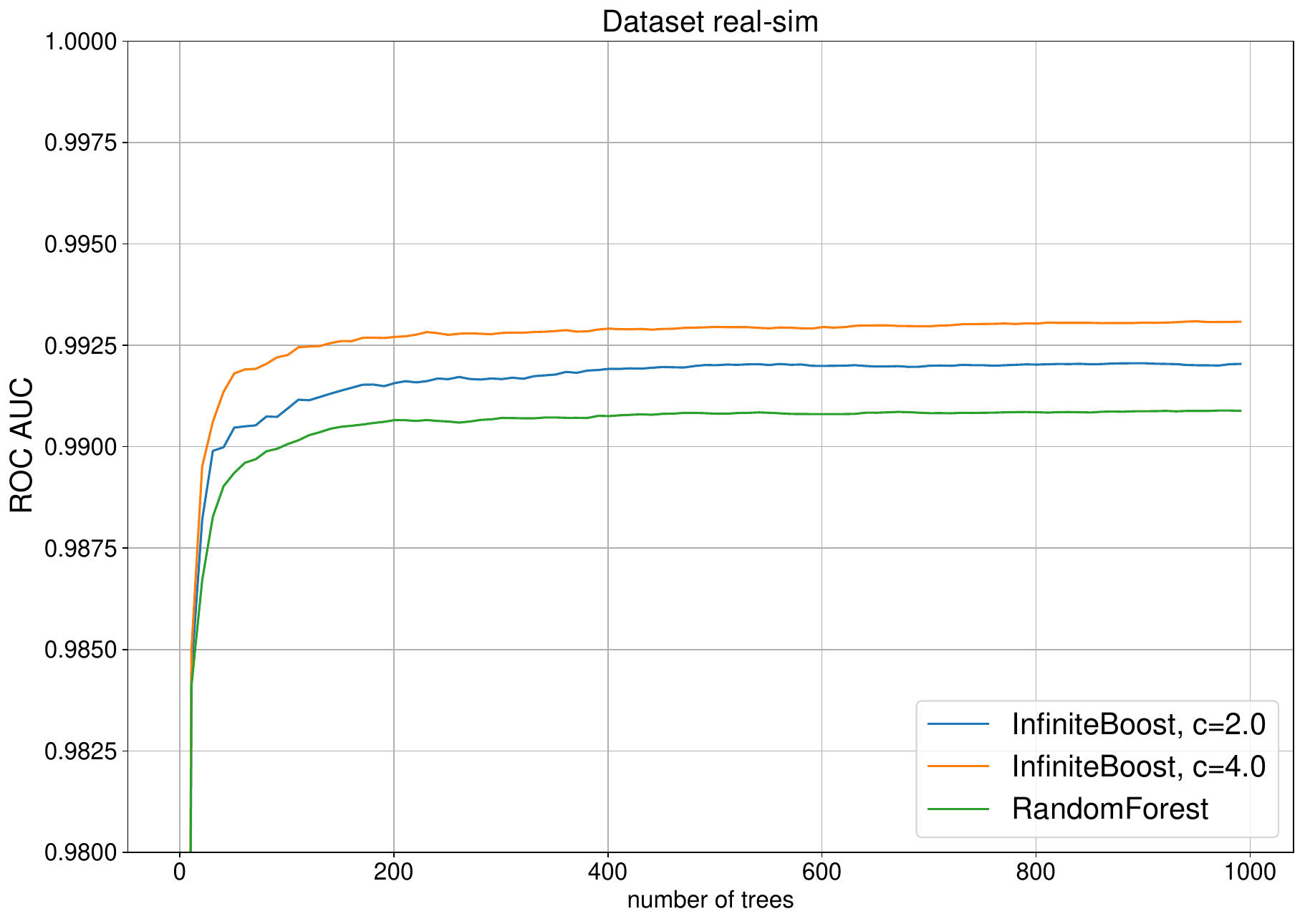}
    \end{multicols}
    \caption{ROC AUC quality on covertype (left) and real-sim (right) datasets for random forest and infinite boosting with capacities $c=2$ and $c=4$. \label{fig:inf}}
  \end{figure*}
\else
  \begin{figure*}[!h]
    \centering
      \includegraphics[width=0.5\linewidth]{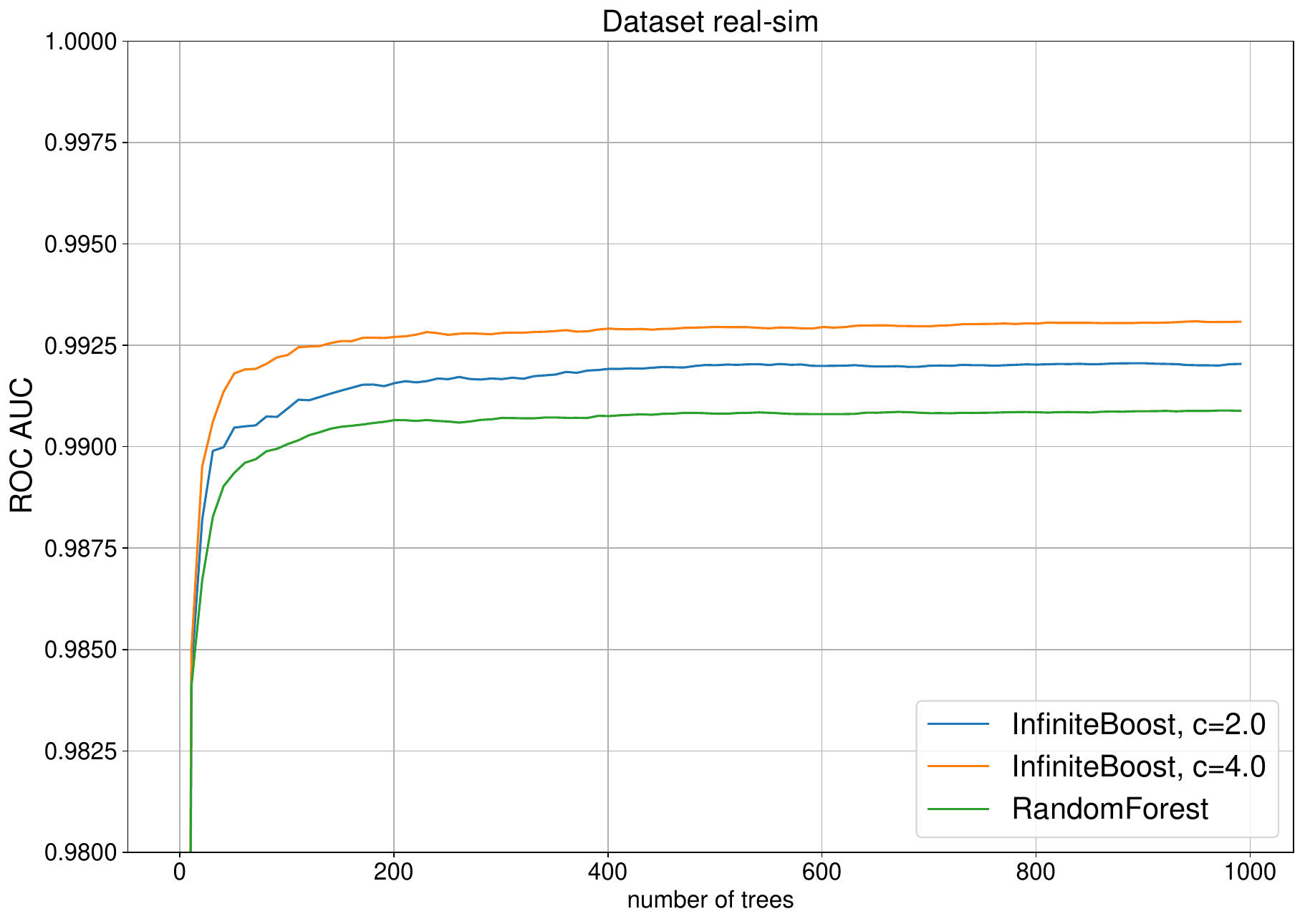}
      \caption{ROC AUC quality on real-sim dataset for random forest and infinite boosting with capacities $c=2$ and $c=4$. \label{fig:inf}}
  \end{figure*}
\fi

\subsection{InfiniteBoost and gradient boosting}
Datasets used for experiments are listed in Table~\ref{tab:gb-data}. 
For Higgs dataset 1 million samples are used for training and 500,000 samples are used for test. 
For YearPredictionMSD dataset random 75\% are taken as training and the remaining samples are used for test purposes.
InifiniteBoost uses adaptive capacity for classification and regression problems.
For ranking task the fixed capacity value is used because the loss function is not convex in this case and the adaptation on the holdout significantly underestimates capacity.
For all tasks the same hyperparameters are used for gradient boosting and InfiniteBoost: subsample is set to 0.7, max features --- 0.7, max depth --- 7. 
For gradient boosting shrinkage is varied to compare with InfiniteBoost.
Other hyperparameters settings are tested and similar behavior is observed.
In Figures~\ref{fig:gb-auc}, \ref{fig:gb-mse}, \ref{fig:gb-ndcg} the learning curves are presented.
It can be seen that InifiniteBoost provides similar quality as gradient boosting (with appropriate shrinkage) and at the same time it is free from over-fitting effect. 
As a number of trees increases, the quality of InfiniteBoost tends to some constant, which confirms theorems.
Worthnoty, found optimal capacities differ significantly for trees of different complexity (Figure~\ref{fig:gb-depth}).

\insertplotsgbnips

\subsection{InfiniteBoost and random forest}

Random forests are known for providing good results with standard hyperparameters.
InfiniteBoost with small capacity almost does not encounter mistakes done at the previous stages,
thus, behaves similarly to a random forest.

In comparison we use an implementation of random forest from scikit-learn~\cite{key-sklearn}. 
Datasets (see Table~\ref{tab:rf-data}) with known superior performance of random forest were taken.
To have side-by-side comparison InfiniteBoost uses deep trees with the same parameters as random forest.
Such deep trees over-fit to the training data, therefore, they are usually considered to be inappropriate for boosting.
In the experiments InfiniteBoost is tested with different fixed capacity values and it is observed that adding a bit of boosting behavior to random forest by setting a small capacity improves the model for 2 of 3 datasets (see Table~\ref{tab:rf-auc}).
Proposed algorithm is capable of using the mistakes made by previous trees on out-of-bag samples, which makes boosting possible.
In experiments with large number of trees, quality of InfiniteBoost converges, showing a behavior similar to random forest (Figure~\ref{fig:inf}).

\insertplotsforestnips

\clearpagebiblio 
\section{Related works}
There are different attempts of combining properties of random forests and gradient boosting.
Bagging methods are mainly proposed to effectively reduce the variance of regression predictors, while they leave bias relatively unchanged. 
To reduce both bias and variance iterated bagging was developed~\cite{key-iterated-bagging}. 
This iterative algorithm trains a sequence of bagging estimators: outcomes of each bagging trained are used to alter the target values for next stages; process repeated until a simple rule stops the process. 
The idea behind bagging is to train each predictor in the ensemble iteratively on the difference between the target and the prediction of the ensemble constructed by this moment (also known as the residual).
Thus, iterated bagging includes the property of gradient boosting: each new predictor in the ensemble accounts for the performance of previous predictors.
Bagging procedure and its out-of-bag estimation are used to obtain an unbiased estimate of the “true” residual. 

Another approach is stochastic gradient boosting~\cite{key-sgb} that introduces randomness into gradient boosting by providing lower correlation between predictors similarly to random forests. 
On each boosting iteration subsample of the training data drawn at random (without replacement) from the full training set is used to fit a new predictor in the ensemble. 
The lower size of subsample the more random samples will differ and the more randomness will be introduced into boosting procedure.
At the same time lower size of subsample reduces the amount of data used for building a new predictor fitting, thus increases the variance of individual predictors.
This version of gradient boosting is known to provide good results and has widest usage in practice.

As discussed above, over-fitting is a well-known problem for gradient boosting. 
An idea of another boosting-bagging hybrid algorithm~\cite{key-dart}, called DART, is to fight this issue by employing dropouts idea for ensembles of trees: muting complete trees as opposed to muting features in random forests. 
During each boosting iteration a randomly chosen subset of trees forms a new ensemble~$M'$. 
A new regression tree is fitted to the negative gradient of the loss function with respect to the predictions obtained from the ensemble~$M'$.
Adding a new tree to the initial ensemble is accompanied by a normalization step.
Firstly, the new tree is scaled by a factor $1/k$, where $k$ is the number of dropped trees from the initial ensemble, to provide the same order of magnitude for the new tree as the dropped trees. 
Secondly, the new tree and the dropped trees are scaled by a factor of $k/(k+1)$ and the new tree is added to the ensemble.
The last scaling ensures that the combined effect of the dropped trees together with the new tree remains the
same as the effect of the dropped trees alone before the introduction of the new tree.
InfiniteBoost and DART have similarities in the normalization step procedure, however, DART does not use correction constant (capacity), like InfiniteBoost.

DART implements gradient boosting, if no tree is dropped, and random forests, if all the trees are dropped.
With comparison to gradient boosting, this algorithm has no shrinkage hyperparameter, which is replaced by dropout rate, a fraction of trees muted on each iteration.
However, construction of large ensembles is not feasible in this algorithm from the computational point,
since on each learning iteration it is needed to prepare a new target by predicting the training set with a random subset of already constructed trees.
This leads to the quadratic (not linear as for gradient boosting) dependence between training time and the number of built trees.\footnote{
There is a mode of DART algorithm, called $\epsilon$-algorithm, with dropping only one tree from the ensemble on each iteration.
It is linear in time w.r.t. the number of trees in the ensemble.}
Also, the algorithm does not converge to some point for arbitrarily large number of trees because the contribution of newly-built tree does not tend to zero, which causes over-fitting.

Simple approach of combining random forests and boosting is proposed in~\cite{key-bagboo}, called BagBoo. 
Each predictor in the bootstrap aggregated ensemble is gradient boosting. 
This approach aims to be well parallelized (each gradient boosting predictor contains 10-20 trees).

\section{Conclusion}

We proposed a new hybrid algorithm called InfiniteBoost with the aim of combining positive elements of two algorithms. 
Empirically InfiniteBoost shows the quality not worse than random forests and asymptotically (with increasing the number of trees) not worse than gradient boosting trained with different shrinkage values.
This could save the time spent on shrinkage/size of ensemble tuning for gradient boosting in applications.
Also, experiments demonstrate that learning curve on the unseen data for InfiniteBoost tends to the constant starting from some iteration of boosting, a favorable property of random forests, which we theoretically proved for InfiniteBoost.

\clearpagebiblio

\end{document}